\documentclass[sigconf]{acmart}
\usepackage{enumitem}
\usepackage[htt]{hyphenat}
\usepackage{listings}
\usepackage{courier} 
\usepackage{tikz}
\usepackage{multirow} 
\usetikzlibrary{positioning, shapes, arrows.meta, calc}
\AtBeginDocument{%
	}

\setcopyright{acmlicensed}
\copyrightyear{2018}
\acmYear{2018}
\acmDOI{XXXXXXX.XXXXXXX}
\acmConference[Conference acronym 'XX]{Make sure to enter the correct
  conference title from your rights confirmation email}{June 03--05,
  2018}{Woodstock, NY}
\acmISBN{978-1-4503-XXXX-X/2018/06}

\begin{document}
	
	\title{PPoGA: Predictive Plan-on-Graph with Action for Knowledge Graph Question Answering}
	
	%
	\author{MinGyu Jeon}
	\affiliation{%
		\institution{MODULABS}
		\city{}
		\country{}
	}
	\email{jkmcoma7@gmail.com}

	\author{SuWan Cho}
	\affiliation{%
		\institution{MODULABS}
		\city{}
		\country{}
	}
	\email{swannyy7@gmail.com}

	\author{JaeYoung Shu}
	\affiliation{%
		\institution{MODULABS}
		\city{}
		\country{}
	}
	\email{tjwodud04@gmail.com}

	
	\renewcommand{\shortauthors}{Jeon et al.}
	
	\begin{abstract}
Large Language Models (LLMs) augmented with Knowledge Graphs (KGs) have advanced complex question answering, yet they often remain susceptible to failure when their initial high-level reasoning plan is flawed. This limitation, analogous to cognitive functional fixedness, prevents agents from restructuring their approach, leading them to pursue unworkable solutions. To address this, we propose \textbf{PPoGA (Predictive Plan-on-Graph with Action)}, a novel KGQA framework inspired by human cognitive control and problem-solving. PPoGA incorporates a \textbf{Planner-Executor} architecture to separate high-level strategy from low-level execution and leverages a \textbf{Predictive Processing} mechanism to anticipate outcomes. The core innovation of our work is a self-correction mechanism that empowers the agent to perform not only \textbf{Path Correction} for local execution errors but also \textbf{Plan Correction} by identifying, discarding, and reformulating the entire plan when it proves ineffective. We conduct extensive experiments on three challenging multi-hop KGQA benchmarks: \textbf{GrailQA}, \textbf{CWQ}, and \textbf{WebQSP}. The results demonstrate that PPoGA achieves state-of-the-art performance, significantly outperforming existing methods. Our work highlights the critical importance of metacognitive abilities like problem restructuring for building more robust and flexible AI reasoning systems.
	\end{abstract}




\maketitle
	\section{Introduction}

Large Language Models (LLMs) have recently demonstrated unprecedented success in few-shot and zero-shot generalization, showcasing their potential to solve complex reasoning tasks \cite{brown2020language, chowdhery2022palm, wei2022chain, ouyang2022training}. However, LLMs face inherent limitations such as the computational opacity of their internal knowledge, a lack of up-to-date information, and the phenomenon of hallucination, where they generate factually incorrect information \cite{chen2024plan, sun2023reasoning, sun2023thinking, ji2023survey}. To mitigate these limitations, the KG-augmented LLM paradigm has emerged as a promising solution, integrating Knowledge Graphs (KGs)---which provide explicit and editable factual knowledge---into the reasoning process of LLMs \cite{sun2023reasoning, sun2023thinking, pan2024unifying, yasunaga2021qa, baek2023knowledge}.

Early studies on KG-based reasoning primarily extended the Chain-of-Thought (CoT) to explore reasoning paths within the KG. However, this approach relied solely on the internal reasoning of LLMs, leading to the generation of inaccurate paths that diverge from the actual KG structure \cite{wei2022chain, press2022measuring, creswell2022faithful}. Addressing this issue, Plan-on-Graph (PoG) introduced an advanced methodology that decomposes a question into multiple sub-goals, dynamically explores reasoning paths within the KG, and self-corrects its path upon encountering errors \cite{chen2024plan}.

However, PoG's self-correction mechanism has a fundamental limitation: it is confined to \textbf{analytical solving}, where an efficient solution is sought within a given framework \cite{wertheimer1959productive, duncker1945problem}. If the initial high-level plan---the "framework" itself---is unsuitable for solving the problem, the agent can become \textbf{functionally fixed} on that plan, failing to redefine the problem space and ultimately leading to failure \cite{duncker1945problem, newell1972human}.

To overcome this limitation, this paper proposes \textbf{PPoGA (Predictive Plan-on-Graph with Action)}, a novel framework that computationally models the multi-layered and flexible problem-solving methods of humans. PPoGA begins by mimicking the \textbf{Hierarchical Control of Action} in humans, explicitly separating roles into a \textbf{Planner}, which formulates high-level strategies, and an \textbf{Executor}, which carries out specific procedures. This division reduces the \textbf{cognitive load} on each component, thereby maximizing the efficiency and stability of the entire system \cite{wang2024plan, norman1986attention, kahneman2011thinking}. Building on this structure, PPoGA incorporates a \textbf{Predictive Processing} mechanism, inspired by how the brain learns from \textbf{prediction errors}, to deepen its reasoning by simulating the outcomes of actions before they are taken \cite{clark2013predictive, friston2010free}.

Furthermore, the core contribution of PPoGA lies in its self-correction mechanism, which emulates the dual human ability to find solutions within a problem's framework and to \textbf{restructure} the problem when the framework itself is flawed \cite{wertheimer1959productive}. Through this mechanism, PPoGA performs both \textbf{Path Correction}---optimizing paths within a given plan---and \textbf{Plan Correction}---recognizing the failure of the plan itself and formulating a new strategy. This enables PPoGA to achieve a higher level of robust and flexible problem-solving capabilities compared to previous studies \cite{flavell1979metacognition}.  The key contributions of this paper are as follows:

\begin{itemize}[topsep=5pt, partopsep=5pt]
    \item First, we propose \textbf{PPoGA}, a novel Knowledge Graph Question Answering (KGQA) framework that integrates a Planner-Executor role separation, prediction-based reasoning, and dynamic \textbf{Path Correction} into a single, coherent architecture.
    \item Second, we are the first to propose and implement a self-correction mechanism that allows an agent to perform both \textbf{Path Correction} for execution-level errors and \textbf{Plan Correction} by recognizing plan failure and reformulating strategy.
    \item Third, through experiments on complex KGQA benchmarks, we demonstrate that \textbf{PPoGA} achieves higher accuracy and robustness than existing static plan-based agents.
\end{itemize}

\section{Methodology}
\label{sec:methodology}

In this section, we elucidate the architecture and operational workflow of our proposed framework, \textbf{PPoGA (Predictive Plan-on-Graph with Action)}. Our method is predicated on the cognitive principle of Hierarchical Control of Action, which separates high-level strategic reasoning from low-level environmental interaction. This is realized through a dual-component architecture comprising a \textbf{Planner} and an \textbf{Executor}, whose operations are coordinated via a unified, three-tiered \textbf{Integrated Memory Architecture}. The interplay between these components enables a dynamic, multi-stage workflow designed for robust and adaptive problem-solving on knowledge graphs.

The \textbf{Planner}, built upon a LLM, serves as the cognitive core of the agent. It is responsible for high-level reasoning tasks, including initial strategy formulation, task decomposition into a multi-step plan, outcome prediction, and self-correction based on environmental feedback. The \textbf{Executor} acts as the interface to the external environment—in this case, a KG. It translates abstract instructions from the Planner into executable SPARQL queries, executes them, and returns structured observations, thereby closing the loop between reasoning and action. All interactions and state information are mediated through the Integrated Memory Architecture, which provides the Planner with comprehensive situational awareness throughout the problem-solving process.

\subsection{The PPoGA Workflow}

\begin{figure*}[t!]
    \centering
    \includegraphics[width=0.8\textwidth]{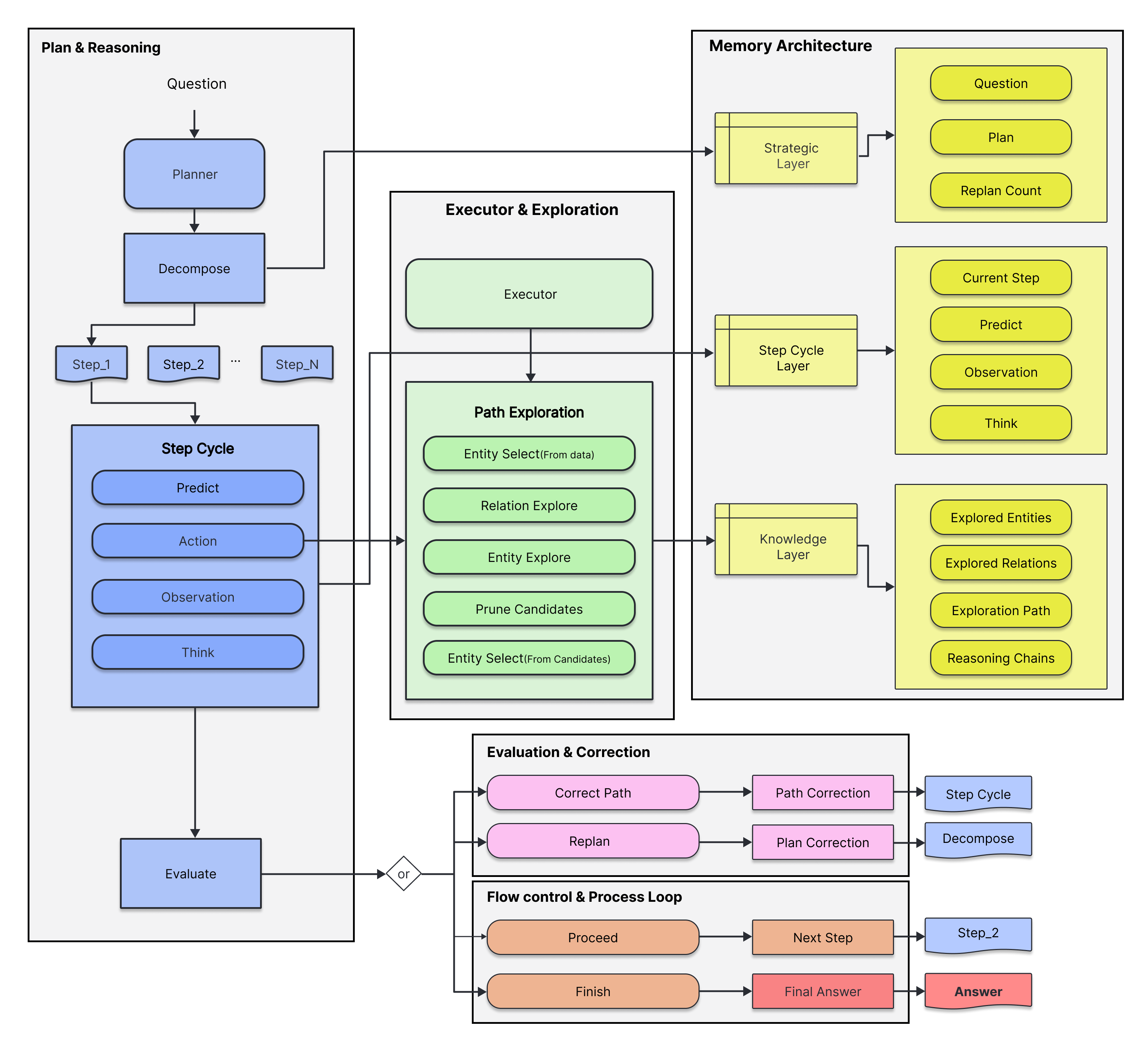}
    \caption{
        \textbf{The PPoGA Workflow and Architecture.} 
        The framework consists of a \textbf{Planner} (reasoning) and an \textbf{Executor} (action), which interact through a three-layered \textbf{Integrated Memory Architecture}. The workflow follows an iterative cycle of Decomposition, a four-stage Step Cycle (Predict, Act, Observe, Think), and Evaluation, enabling both \textbf{Plan} and \textbf{Path Correction}.
    }
    \Description{Block diagram of the PPoGA framework showing the Planner and Executor components interacting via a three-layer Integrated Memory, and the iterative cycle of Predict, Act, Observe, and Think with plan and path correction.}
    \label{fig:ppoga_framework}
\end{figure*}

The PPoGA workflow emulates a structured but flexible human-like reasoning process, beginning with \textbf{Decomposition} and proceeding through an iterative \textbf{Step Cycle} and an \textbf{Evaluation} phase, as illustrated in Figure~\ref{fig:ppoga_framework}.

Upon receiving a question, the Planner first performs \textbf{Decomposition}, creating a multi-step strategic plan to find the answer. This initial plan is a dynamic blueprint composed of discrete steps, where each step encapsulates a unique objective, a description, and a status tracker (e.g., \textit{not started, in progress, completed}).

Once the plan is formulated, the agent executes it sequentially via an iterative, four-stage \textbf{Step Cycle}: \textbf{Predict, Act, Observe, and Think}. The cycle begins with \textbf{Prediction}, where the Planner simulates the expected outcome of the current step's objective before any action is taken. This proactive mechanism allows the agent to anticipate and identify potential deviations. Following prediction, the \textbf{Action} phase, termed \textbf{Path Exploration}, is executed by the Executor to retrieve information from the KG. This process starts by identifying the topic entity from the user's query. The Executor then explores the KG by retrieving all connected relations and subsequent entities to form a set of candidate triples. To manage combinatorial explosion, if the number of candidates exceeds a predefined threshold (e.g., 70), an embedding-based pruning mechanism is employed to retain only the most semantically relevant candidates. From this pruned set, the Executor selects the single most promising entity, updating the agent's memory with the choice and its corresponding rationale. The result of this action forms the \textbf{Observation}, which is then compared against the initial prediction to compute an error signal. This comparison informs the \textbf{Think} stage, where the Planner generates an internal reflection on the outcome, leading to a decision in the final phase.

The culmination of each Step Cycle is the \textbf{Evaluation and Self-Correction} phase, where the Planner determines the next course of action. If the step's objective is met, the agent \textbf{proceeds} to the next step in the plan. If the action was unsuccessful but the overall plan remains sound, it initiates \textbf{Path Correction}, re-attempting the current step while leveraging memory to avoid previously failed exploration paths. Should repeated attempts fail, the agent concludes the plan itself may be flawed and triggers \textbf{Plan Correction (Replan)}, returning to the decomposition phase to formulate a new strategy based on the knowledge gathered thus far. The workflow \textbf{finishes} when the agent determines it has sufficient information to synthesize a final answer.

\subsection{Integrated Memory Architecture}

The entire workflow is underpinned by the Integrated Memory Architecture, a centralized cognitive workspace structured into three interconnected layers that operate at different levels of abstraction. The \textbf{Strategic Layer} maintains the highest-level context, storing the initial question, the overall plan, and metadata such as a replan counter to prevent infinite loops; it provides the foundational knowledge required for Plan Correction. Operating at the level of a single action, the \textbf{Step Cycle Layer} holds transient state information like the current step's objective, predictions, and observations, which is crucial for enabling \textbf{Path Correction}. Finally, the \textbf{Knowledge Layer} serves as the persistent repository for all factual information gathered from the KG, accumulating explored entities, relations, and reasoning chains to ensure that knowledge is retained and redundant actions are avoided.

	\section{Experiment}

\subsection{Experimental Setup}

We demonstrate the effectiveness of the proposed framework, \textbf{PPoGA (Predictive Plan-on-Graph with Action)}, on the well-established multi-hop KGQA benchmarks \textbf{GrailQA} \cite{gu2021beyond}, \textbf{CWQ} \cite{talmor2018web}, and \textbf{WebQSP} \cite{yih2016value}.
The experimental settings are consistent with previous studies such as ToG and PoG \cite{sun2023thinking, chen2024plan}, and the same testing samples are utilized for computational efficiency.

As in previous KGQA literature \cite{gu2021beyond, jiang2023unikgqa, luo2024reasoning}, we use \textbf{Exact Match Accuracy (Hits@1)} as the evaluation metric \cite{li2023kbbinder, wang2023knowledgedriven}.
This criterion measures whether the entity predicted by the model exactly matches the ground truth answer \cite{gu2021beyond, sun2023thinking}.

The comparison baselines follow PoG \cite{chen2024plan} and can be classified into categories:
\begin{itemize}[topsep=3pt, partopsep=3pt]
    \item \textbf{LLM-only:} a conventional language model without KG integration \cite{brown2020language, wei2022chain, wang2023selfconsistency, zhou2023least}.
    \item \textbf{Fine-tuned KG-augmented LLM:} a model fine-tuned on KGQA datasets with parameter updates \cite{jiang2023unikgqa, shu2022tiara, luo2024reasoning, yu2023decaf, zhang2023fckbqa, ye2022rngkbqa}.
    \item \textbf{Prompting KG-augmented LLM (GPT-3.5):} a prompting-based reasoning model using GPT-3.5 \cite{li2023kbbinder, wang2023knowledgedriven, jiang2023structgpt, xiong2024interactive, sun2023thinking}.
\end{itemize}

\subsection{Results}

We evaluate PPoGA across three standard benchmarks.
The \textbf{GrailQA} benchmark measures the ability of models to generalize under four evaluation settings: \textit{Overall}, \textit{I.I.D.}, \textit{Compositional}, and \textit{Zero-Shot} \cite{gu2021beyond}.
These configurations are designed to evaluate, respectively, overall performance, in-domain reasoning, compositional generalization, and out-of-distribution reasoning capability \cite{chen2024plan, sun2023thinking}.
Additionally, we assess performance on \textbf{Complex WebQuestions (CWQ)} \cite{talmor2018web} and \textbf{WebQuestionsSP (WebQSP)} \cite{yih2016value} to validate the framework's effectiveness on other complex question-answering datasets.

The main results are presented in Table~\ref{tab:grailqa_comparison} (for GrailQA) and Table~\ref{tab:cwq_webqsp_comparison} (for CWQ and WebQSP).
Our experiments are conducted using GPT-3.5 to ensure a fair and direct comparison against the prompting-based baselines in the same category.
PPoGA consistently achieves state-of-the-art performance among all \textit{Prompting KG-Augmented LLM (GPT-3.5)} baselines across all three datasets.

Most notably, PPoGA demonstrates a clear and consistent performance improvement over its direct predecessor, PoG \cite{chen2024plan}.
On the \textbf{GrailQA} benchmark (Table~\ref{tab:grailqa_comparison}), PPoGA achieves an overall accuracy of 78.9, surpassing PoG's 76.5. This trend holds across the generalization splits, especially in the challenging \textit{Zero-shot} split (82.3 vs. 81.7).
This superiority is further confirmed on \textbf{CWQ} and \textbf{WebQSP} (Table~\ref{tab:cwq_webqsp_comparison}), where PPoGA (64.5 on CWQ, 83.1 on WebQSP) outperforms PoG

\begin{table}[!t]
    \centering
    \small
    \setlength{\tabcolsep}{4pt}
    \begin{tabular*}{\columnwidth}{@{\extracolsep{\fill}}lcccc}
            \toprule
            \textbf{Method} & \textbf{Overall} & \textbf{I.I.D.} & \textbf{Compositional} & \textbf{Zero-shot} \\
            \midrule
            \multicolumn{5}{c}{\textit{LLM-Only}} \\
            \midrule
            IO Prompt~\cite{brown2020language} & 29.4 & - & - & - \\
            CoT~\cite{wei2022chain} & 28.1 & - & - & - \\
            SC~\cite{wang2023selfconsistency} & 29.6 & - & - & - \\
            \midrule
            \multicolumn{5}{c}{\textit{Fine-Tuned KG-Augmented LLM}} \\
            \midrule
            RnG-KBQA~\cite{ye2022rngkbqa} & 68.8 & 86.2 & 63.8 & 63.0 \\
            TIARA~\cite{shu2022tiara} & 73.0 & 87.8 & 69.2 & 68.0 \\
            FC-KBQA~\cite{zhang2023fckbqa} & 73.2 & 88.5 & 70.0 & 67.6 \\
            Pangu~\cite{gu2023pangu} & 75.4 & 84.4 & 74.6 & 71.6 \\
            FlexKBQA~\cite{li2024flexkbqa} & 62.8 & 71.3 & 59.1 & 60.6 \\
            GAIN~\cite{shu2024gain} & 76.3 & 88.5 & 73.7 & 71.8 \\
            \midrule
            \multicolumn{5}{c}{\textit{Prompting KG-Augmented LLM (GPT-3.5)}} \\
            \midrule
            KB-BINDER~\cite{li2023kbbinder} & 50.6 & - & - & - \\
            ToG~\cite{sun2023thinking} & 68.7 & 70.1 & 56.1 & 72.7 \\
            PoG~\cite{chen2024plan} & 76.5 & 76.3 & 62.1 & 81.7 \\
            \textbf{PPoGA (ours)} & \textbf{78.9} & \textbf{78.6} & \textbf{64.7} & \textbf{82.3} \\
            \bottomrule
    \end{tabular*}
    
    \caption{Performance comparison of different methods on GrailQA.}
    \label{tab:grailqa_comparison}
\end{table}

\begin{table}[!t]
    \centering
    \small
    \setlength{\tabcolsep}{4pt}
    \begin{tabular*}{\columnwidth}{@{\extracolsep{\fill}}lcc}
            \toprule
            \textbf{Method} & \textbf{CWQ} & \textbf{WebQSP} \\
            
            \midrule
            \multicolumn{3}{c}{\textit{LLM-Only}} \\
            \midrule
            IO Prompt~\cite{brown2020language} & 37.6 & 63.3 \\
            CoT~\cite{wei2022chain} & 38.8 & 62.2 \\
            SC~\cite{wang2023selfconsistency} & 45.4 & 61.1 \\
            
            \midrule
            \multicolumn{3}{c}{\textit{Fine-Tuned KG-Augmented LLM}} \\
            \midrule
            UniKGQA~\cite{jiang2023unikgqa} & 51.2 & 79.1 \\
            TIARA~\cite{shu2022tiara} & - & 75.2 \\ 
            RE-KBQA~\cite{cao2023pay} & 50.3 & 74.6 \\
            DeCAF~\cite{yu2023decaf} & 70.4 & 82.1 \\
            ROG~\cite{luo2024reasoning} & 62.6 & 85.7 \\

            \midrule
            \multicolumn{3}{c}{\textit{Prompting KG-Augmented LLM (GPT-3.5)}} \\
            \midrule
            KD-COT~\cite{wang2023knowledgedriven} & 50.5 & 73.7 \\
            KB-BINDER~\cite{li2023kbbinder} & 54.3 & 74.4 \\ 
            StructGPT~\cite{jiang2023structgpt} & 57.1 & 72.6 \\
            ToG~\cite{sun2023thinking} & - & 76.2 \\ 
            PoG~\cite{chen2024plan} & 63.2 & 82.0 \\
            \textbf{PPoGA (ours)} & \textbf{64.5} & \textbf{83.1} \\
            
            \bottomrule
        \end{tabular*}
    
    \caption{Performance comparison (Hits@1) on CWQ and WebQSP, aligned with PoG baselines.}
    \label{tab:cwq_webqsp_comparison}
\end{table}

	\section{Conclusion}

In this paper, we introduced \textbf{PPoGA (Predictive Plan-on-Graph with Action)}, a novel framework for complex KGQA that addresses the critical limitation of functional fixedness in existing planning-based agents.
By computationally modeling principles of human cognition, PPoGA moves beyond simple error correction to enable true problem restructuring.
Our core contributions are the explicit separation of high-level reasoning and low-level execution through a \textbf{Planner-Executor} architecture, the integration of a \textbf{Predictive Processing} mechanism for proactive error anticipation, and, most importantly, the introduction of a \textbf{Two-Level Self-Correction} mechanism.
This dual-correction system empowers the agent to perform not only \textbf{Path Corrections} at the execution level but also \textbf{Plan Corrections} by identifying and abandoning flawed high-level plans.

Our experimental results on three challenging KGQA benchmarks empirically validate the effectiveness of our approach.
PPoGA consistently outperforms existing state-of-the-art methods, demonstrating that its ability to flexibly adapt and reformulate strategies leads to superior accuracy and robustness.
These findings underscore the value of building agents that can reason about their own reasoning process, a key aspect of metacognition.

The success of PPoGA suggests a promising direction for developing more autonomous and resilient AI systems.
Future work could explore applying this hierarchical and self-correcting cognitive architecture to other complex, multi-step reasoning domains such as embodied AI, software engineering, and scientific discovery, where the ability to dynamically restructure a problem is paramount.

\begin{acks}
This research was supported by Brian Impact Foundation, a non-profit organization dedicated to the advancement of science and technology for all.
\end{acks}

	\bibliographystyle{ACM-Reference-Format}
	\bibliography{references}

\end{document}